# A Novel Fuzzy Logic Based Adaptive Super-twisting Sliding Mode Control Algorithm for Dynamic Uncertain Systems

Abdul Kareem[1] and Dr. Mohammad Fazle Azeem[2]

[1] Research Scholar, Department of Electronics & Communication Engineering, St. Peter's University, Chennai, India.
Associate Professor, Department of Electronics & Communication Engineering, Sahyadri College of Engineering and Management, Mangalore, India.
afthabakareem@gmail.com
[2] PA College of Engineering, Mangalore, Karnataka, India.
mf.azeem@gmail.com

## ABSTRACT

*This paper presents a novel fuzzy logic based Adaptive Super-twisting Sliding Mode Controller for the control of dynamic uncertain systems. The proposed controller combines the advantages of Second order Sliding Mode Control, Fuzzy Logic Control and Adaptive Control. The reaching conditions, stability and robustness of the system with the proposed controller are guaranteed. In addition, the proposed controller is well suited for simple design and implementation. The effectiveness of the proposed controller over the first order Sliding Mode Fuzzy Logic controller is illustrated by Matlab based simulations performed on a DC-DC Buck converter. Based on this comparison, the proposed controller is shown to obtain the desired transient response without causing chattering and error under steady-state conditions. The proposed controller is able to give robust performance in terms of rejection to input voltage variations and load variations.*

## 1. INTRODUCTION

Sliding mode control is a powerful control method that can produce a very robust closed-loop system under plant uncertainties and external disturbances [1]-[3], because the sliding mode can be designed entirely independent of these effects. Also, Sliding mode controllers are inherently stable. However several disadvantages exist for sliding mode control. An assumption for sliding mode control is that the control can be switched from one value to another infinitely fast. In practice, it is impossible to change the control infinitely fast because of the time delay for control computations and physical limitations of switching devices. As a result, chattering occurs in steady state and appears as an oscillation that may excite unmodeled high-frequency dynamics in the system. Hysteresis can be used to control the switching frequency, but a constant switching frequency cannot be guaranteed. However, there is always chattering in the sliding mode when hysteresis is employed. As a result, the system is able to approach the sliding mode but not able to stay on it [1].

The second order sliding-mode algorithm is a good approach to chattering alleviation. But second order sliding mode algorithm requires the knowledge of the values of the derivatives and the knowledge of the perturbation [1].

Super-twisting Sliding Mode algorithm (STA) is a second order Sliding Mode Control algorithm which is a unique absolutely continuous Sliding Mode algorithm, ensuring all the main properties of first order Sliding Mode control for the systems with Lipschitz matched uncertainties with bounded gradients and eliminates the chattering phenomenon. Super-twisting algorithm does not require the knowledge of the values of the derivatives and the knowledge of the perturbation. Also Super-twisting Sliding mode controllers are inherently stable.

Fuzzy Logic control (FLC) has also been applied to control dynamic uncertain systems [2]-[12]. Fuzzy controllers are very suitable for nonlinear time-variant systems and do not need an exact mathematical model for the system being controlled. They are usually designed based on expert knowledge of the system and has become a good approach to overcome the difficulties in finding mathematical models of systems with complex dynamics and unexpected disturbances. The disadvantage is that extensive tuning based on a trial and error method is required for the design. This tuning can be quite time consuming. In addition, the response of system with a fuzzy controller is not easy to predict [3].

Presented in this paper is the application of fuzzy logic in the design of an adaptive Super-twisting Sliding Mode controller for dynamic uncertain systems with an example to regulate the output voltage of a buck converter. The proposed controller combines the advantages of fuzzy logic, Super-twisting Sliding Mode Control and adaptive controllers and has its own unique advantages that facilitate its design and implementation.

## 2. FUZZY LOGIC CONTROL

Though the fuzzy controllers are highly customizable and vary a great degree on various counts, it is often possible to describe certain basic components which inevitably find place in any fuzzy control scheme. The basic components of a generalized fuzzy logic controller are as shown in the Fig 1.

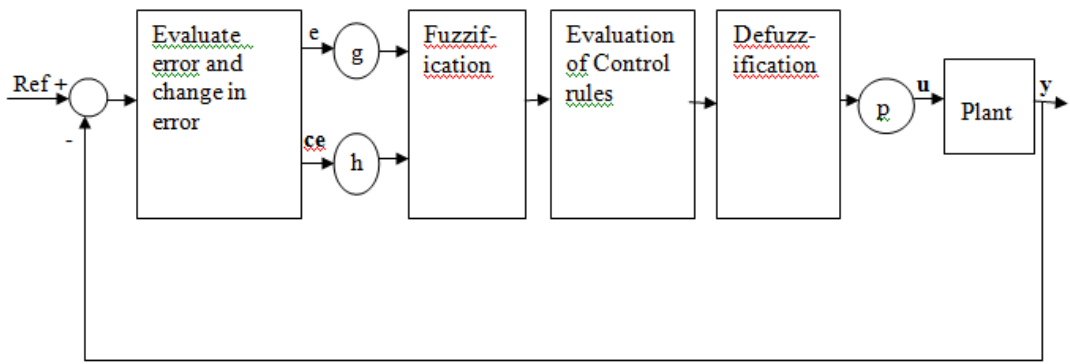

Fig 1. Basic Components of Fuzzy Inference System.

There are two important links between the fuzzy controller and the process or object under control, namely the input and output links. The inputs and outputs are crisp in nature; in the sense that they are the control and controlled parameters of the process or object under control and constitute ordinary numbers, which are understandable, by anyone.

## 2.1. Fuzzification

The controller cannot decipher this input data directly and hence there arises the need for morphing this data to the form comprehensible to the fuzzy system. For this purpose we need the "FUZZIFICATION" block wherein the crisp inputs are fuzzified and each input is given appropriate membership value. Again, the data required to change the crisp input to the fuzzy input is stored in the "KNOWLEDGE BASE". The stored information in the knowledge base is usually the membership function associated with various linguistic variables and the rules to be fired. The nucleus of the whole system is this knowledge base and it has to be designed with utmost care and requires a lot of expertise in the area into which this controller is being used.

## 2.2. Knowledge Base

This knowledge base can be divided into two more sub-blocks namely the "DATA BASE" and "RULE BASE". The former consist mostly the information required for fuzzifying the crisp inputs and later defuzzyfying the fuzzy outputs to a crisp output. The rule base as the name itself suggests consists of set or a table of rules, which are usually formulated from the expert knowledge accumulated over a period of time.

## 2.3. Fuzzy Inference Engine

Next in line is the Fuzzy Inference Engine, which is nothing but an execution unit. It accepts the fuzzy inputs from the fuzzifier and generates fuzzified outputs after the necessary calculations. The fuzzy inference engine evaluates each rule in the Rule Base based on the fuzzified inputs and if this evaluation results in a non-zero output then such rules are said to be 'FIRED'. Now the inference engine would combine all these outputs in accordance with a pre-specified protocol. It is important to note that not all rules need to be equally important, i.e. each rule can be assigned a weight, which indicates its influence on the final output of the inference engine.

## 2.4. De-Fuzzification

The fuzzy outputs from the inference engine are not useful as it is and they need to be converted to crisp output before we can make any proper use of it. This conversion of fuzzy output to crisp output is defined as Defuzzification.

# 3. SUPER-TWISTING SLIDING MODE CONTROL

Super-twisting Sliding Mode algorithm (STA) is a second order Sliding Mode Control algorithm which is a unique absolutely continuous Sliding Mode algorithm, ensuring all the main properties of first order Sliding Mode control for the systems with Lipschitz matched uncertainties with bounded gradients and eliminates the chattering phenomenon. Super-twisting algorithm does not require the knowledge of the values of the derivatives and the knowledge of the perturbation. The work presented by Moreno and Osorio proposed a quadratic like Lyapunov functions for the Super-twisting Sliding Mode controller, making possible to obtain an explicit relation for the controller design parameters. The Sliding surface in Super-twisting Sliding Mode control for second order systems is expressed as

$$S(x) = \dot{e} + ce \qquad (x \neq 0) \qquad (1)$$

where c > 0, is a strictly positive real constant. Here, $e = r-y$ is the error function, $\dot{e}$ is the derivative of error function, $r$ is the reference input and $y$ is the system output. Equation (1) defines the required system dynamics by using error and derivative of error. The super-twisting sliding mode rule to reach $S=0$ can be given as

$$u = -K_1 \phi_1(S) - \int_0^t K_2 \phi_2(S) \, dt$$

$$\phi_1(S) = |S|^{\frac{1}{2}} Sign(S) \qquad (2)$$

$$\phi_2(S) = \frac{1}{2} Sign(S)$$

# 4. SLIDING SURFACE SLOPE ADJUSTMENT OF SUPER-TWISTING SLIDING MODE CONTROLLERS

In this section, the effect of the sliding surface slope of Super-twisting Sliding Mode Controller on the system performance is discussed.

## 4.1. The effect of the sliding surface slope on the system performance

It is a basic fact that the system performance is sensitive to the sliding surface slope $c$ for Super-twisting Sliding Mode Control (STSMC). For instance, if large values of $c$ are considered then the system will give a fast response in STSMC application due to the large values of the control signal but the system may become unstable. If small values of $c$ are chosen, the system will be more stable but the performance of the system may degrade since the system response will become slower due to small values of the control signal. Thus, determination of an optimum $c$ value for a system is an important problem. If the system parameters are unknown or uncertain, the problem becomes more difficult. This problem may be solved by adjusting the slope of the sliding surface of Super-twisting Sliding Mode Controller continuously in real-time. In this way, the performance of the overall system can be improved with respect to the classical Super-twisting Sliding Mode Control.

In this study, the sliding surface slope is continuously updated by multiplying the predetermined slope value $c$ by a new coefficient factor $k_c$. This new coefficient factor $k_c$ is a function of system error $e$ and a new variable $r_v$ named as normalized acceleration [13]. Before introducing the coefficient factor $k_c$ it will be more appropriate to explain the normalized acceleration $r_v$.

## 4.2. The Normalized Acceleration

The normalized acceleration $r_v(k)$ is defined as

$$r_v(k) = \frac{de(k) - de(k-1)}{de(.)} = \frac{dde(k)}{de(.)} \qquad (3)$$

Here, de(k) is the incremental change in error that is defined as

$$de(k) = e(k) - e(k-1) \qquad (4)$$

and dde(k) is called the acceleration in error and it is given by

$$dde(k) = de(k) - de(k-1) \qquad (5)$$

In equation (3), de(.) is chosen as

$$\left| de\left(k\right) \right| \geq \left| de\left(k-1\right) \right| \qquad de\left(.\right) = de\left(k\right) \text{ If } \qquad de\left(.\right) = de\left(k-1\right) \qquad , \text{ else}$$

(6)

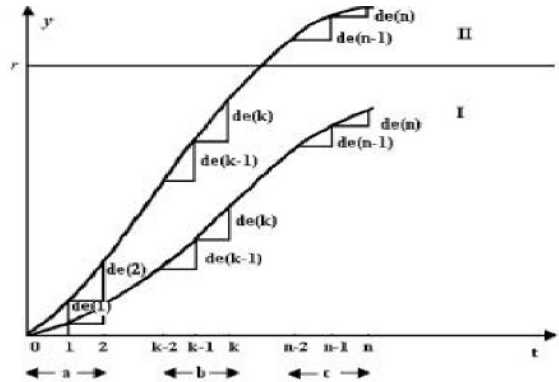

Fig 2. The relative rates of the system responses due to a step input.

When the system response demonstrates a smooth and steady increase or decrease, then the product d$e(k)$.d$e(k-1)$ is positive and "fastness" and "slowness" of the response can be deduced by using this new variable $r_v(k)$. An interesting feature of this normalized acceleration concept is that two system response curves with different time constants can possess the same $r_v(k)$ values as it can be seen from Fig 2. If the absolute value of the change in error |d$e(k)$| is greater than the previous value |d$e(k-1)$| then the system response increases or decreases in a "fast" nature as it is seen in the time interval (a) of Fig 2. Contrary to this case, if the absolute value of the change in error |d$e(k)$| is less than the previous value |d$e(k-1)$| then the system response increases or decreases in a "slow" nature as it is seen in the time interval (c) of Fig 2. Table 1 shows that "fastness" or "slowness" of a system response depends on the signs of both dd$e(k)$ and d$e(.)$. Thus, Equation (3) has been devised in order to normalize the acceleration term dd$e(k)$ while reserving the information about the "fastness" or "slowness" of the system response [13].

The normalized acceleration $r_v(k)$ given in Equation (3) yields us a relative rate information about the system response within a range of [-1,1]. If the system response is very fast, $r_v(k)$ approaches to 1, and if the system response is very slow, it approaches to –1. When the system response increases or decreases with a constant rate as in the time interval (b) of Fig 2, it is considered as a "medium" rate and $r_v(k)$ takes the value of zero [13].

| de(k) or de(k-1) | dde(k) | System response |
|---|---|---|
| Positive | Positive | Fast |
| Positive | Negative | Slow |
| Negative | Positive | Slow |
| Negative | Negative | Fast |

Table 1. Relationship between d$e(.)$ , dd$e(k)$ and the system response.

# 5. THE PROPOSED CONTROL ALGORITHM

The Proposed method is a novel fuzzy logic based adaptive Super-twisting Sliding Mode Control with a simple fuzzy logic based sliding surface slope adjustment of super-twisting sliding mode control algorithm explained in section 3. The sliding surface slope is updated by multiplying the predetermined slope value $c$ by a new coefficient factor $k_c$. This new coefficient factor $k_c$ is generated through the function g($e,r_v$). The metarules for determining $k_c$ through the function g($e,r_v$) can be summarized as follows.

When the system response is not "slow enough"

a) the coefficient factor $k_c$ mainly depends on $e$ and it increases as $e$ increases.
b) it decreases as $r_v$ increases.

When the system response is "slow enough", $k_c$ mainly depends on $r_v$ and it increases very abruptly as $r_v$ value decreases.

## 5.1. Generation of $k_c$ through Fuzzy rules

The meta-rules mentioned above can be formulated in fuzzy terms. Nine linguistic sets {VVS, VS, S, MS, M, ML, L, VL, VVL} are defined for the inputs , $r_v$ and the output $k_c$ . For simplicity, the shapes of all membership functions are chosen as triangular, fuzzy overlapping and symmetric. The scaling factors of inputs are unity. The scaling factor of output $k_c$ can be obtained using trial and error method. The rule table of fuzzy system satisfying the above meta-rules is given in Table 2.

| $e$ / $r_v$ | VVS | VS | S | MS | M | ML | L | VL | VVL |
|---|---|---|---|---|---|---|---|---|---|
| VVS | VVL | S | VS | VS | VS | VS | VS | VVS | VVS |
| VS | VL | S | VS | VS | VS | VS | VS | VS | VVS |
| S | L | MS | VS | VS | VS | VS | VS | VS | VS |
| MS | L | MS | S | VS | VS | VS | VS | VS | VS |
| M | L | MS | S | S | VS | VS | VS | VS | VS |
| ML | L | M | S | S | S | S | S | S | S |
| L | VL | M | S | S | S | S | S | S | S |
| VL | VL | M | MS | S | S | S | S | S | S |
| VVL | VVL | ML | M | MS | MS | MS | MS | MS | MS |

Table 2. Fuzzy Rule Table

## 6. SIMULATIONS

To show the effectiveness of the proposed method, the control of a 20V to 12V DC-DC Buck converter is considered. The performance of the proposed controller is compared with a single input- single output first order Sliding Mode Fuzzy logic Controller (FOSMFLC) with input and output having seven triangular, fuzzy overlapping symmetric membership functions and seven rules. The simulation results are given in Fig 3 to Fig 11.

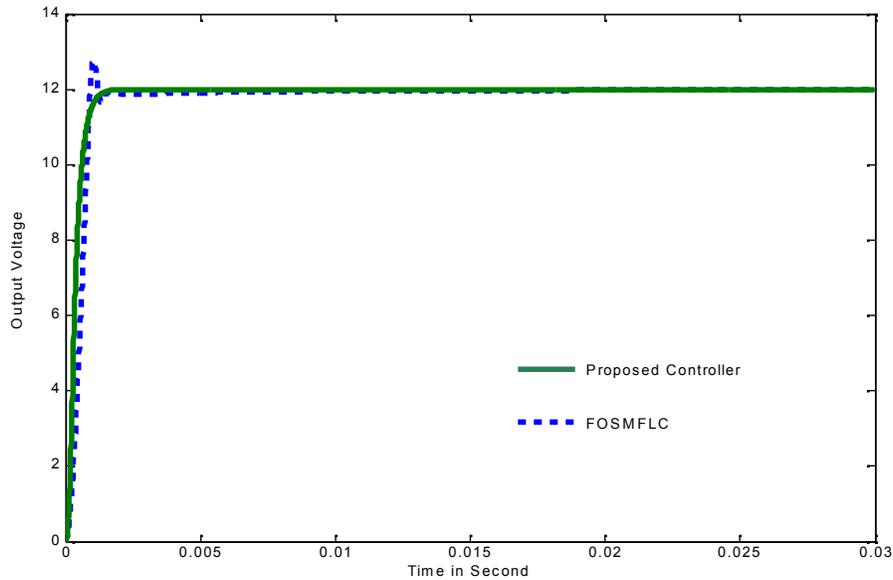

Fig 3. The Proposed Controller and FOSMFLC responses.

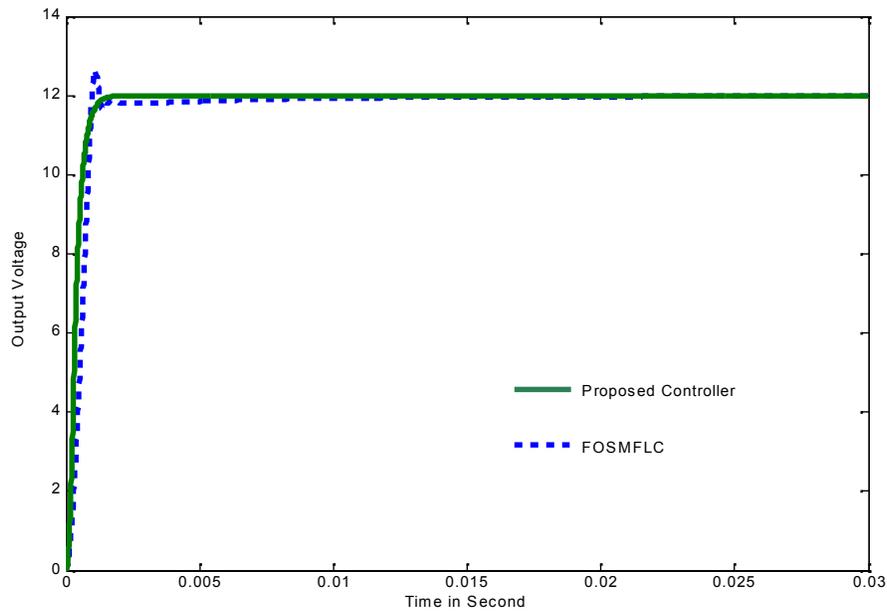

Fig 4. The Proposed Controller and FOSMFLC responses when the input Voltage changes to 18V at Start-up.

27

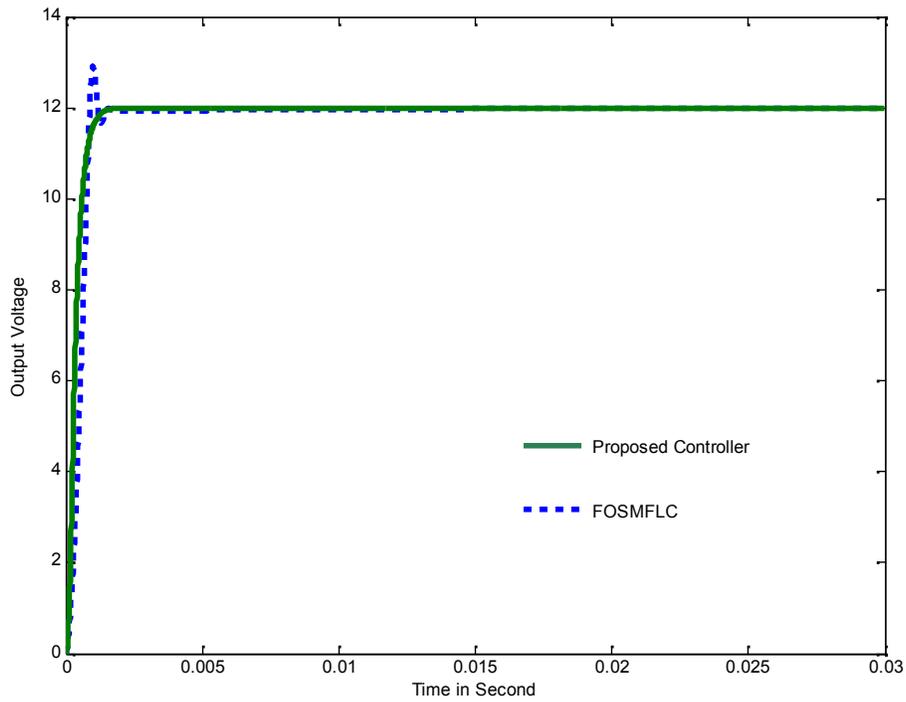

Fig 5. The Proposed Controller and FOSMFLC responses when the input Voltage changes to 22V at Start-up.

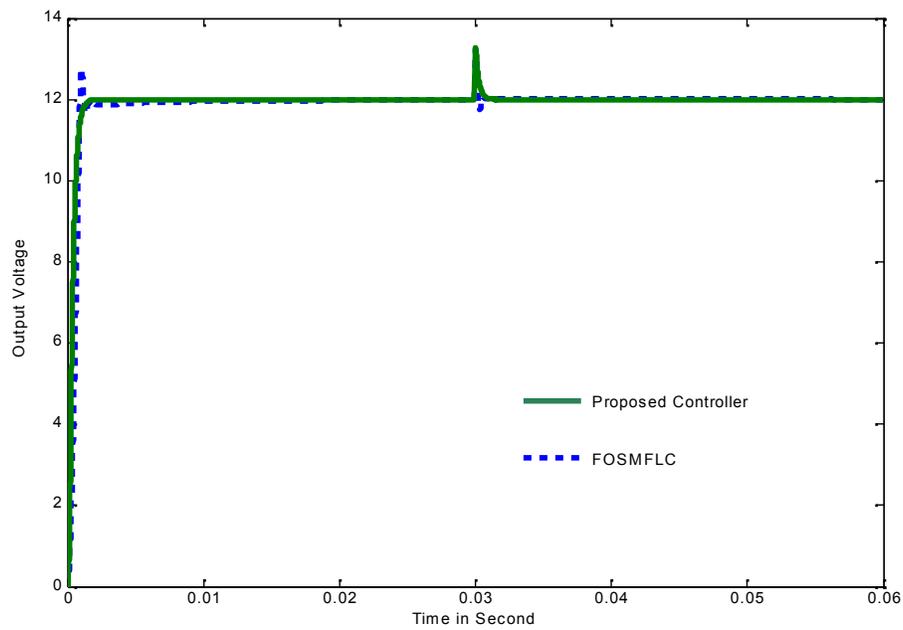

Fig 6. The Proposed Controller and FOSMFLC responses when the input Voltage changes to 22V at Steady-State.

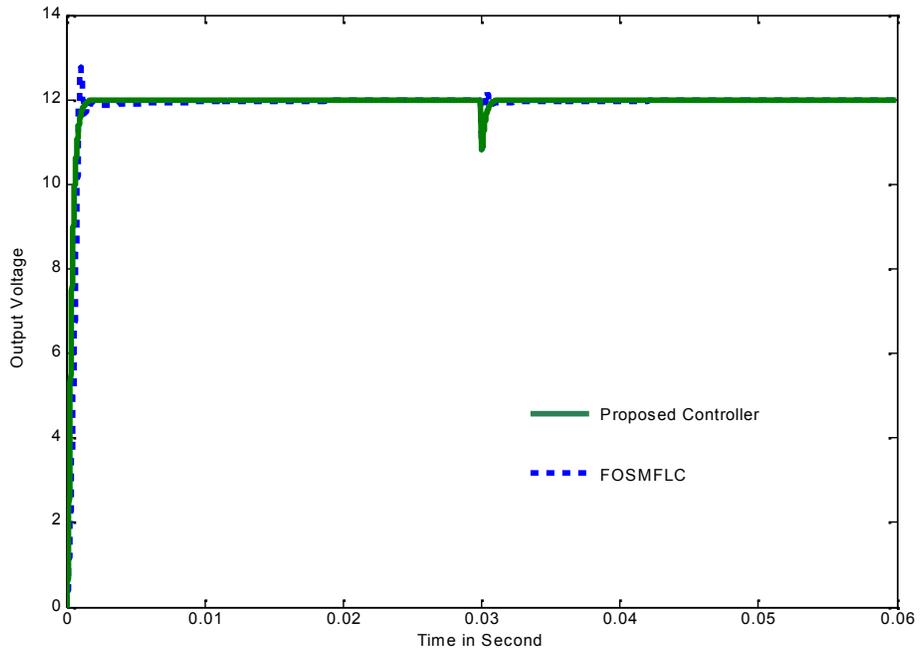

Fig 7. The Proposed Controller and FOSMFLC responses when the input Voltage changes to 18V
at Steady-State.

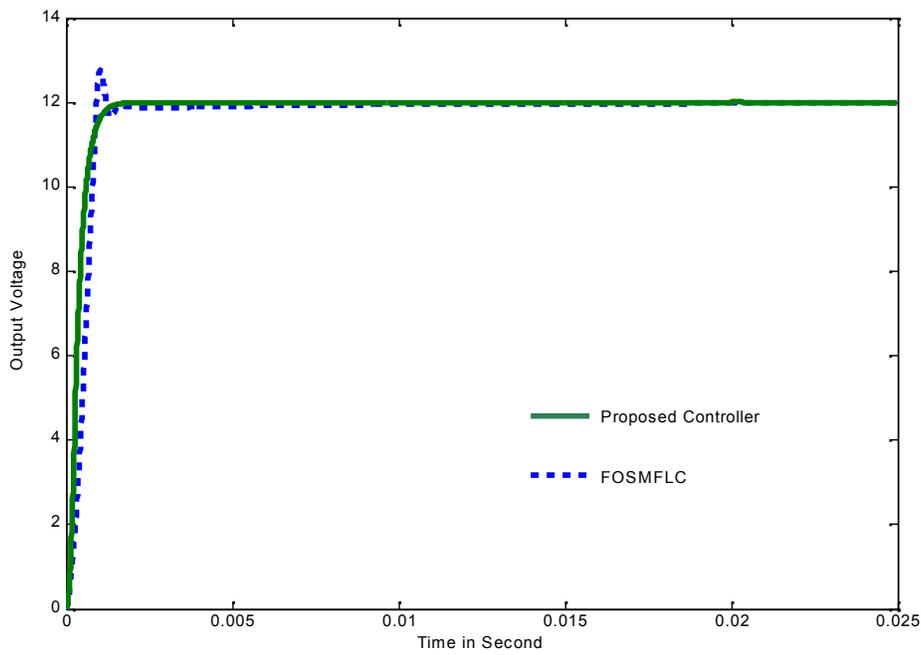

Fig 8. The Proposed Controller and FOSMFLC responses when the load resistance changes from
10Ω to 20Ω.

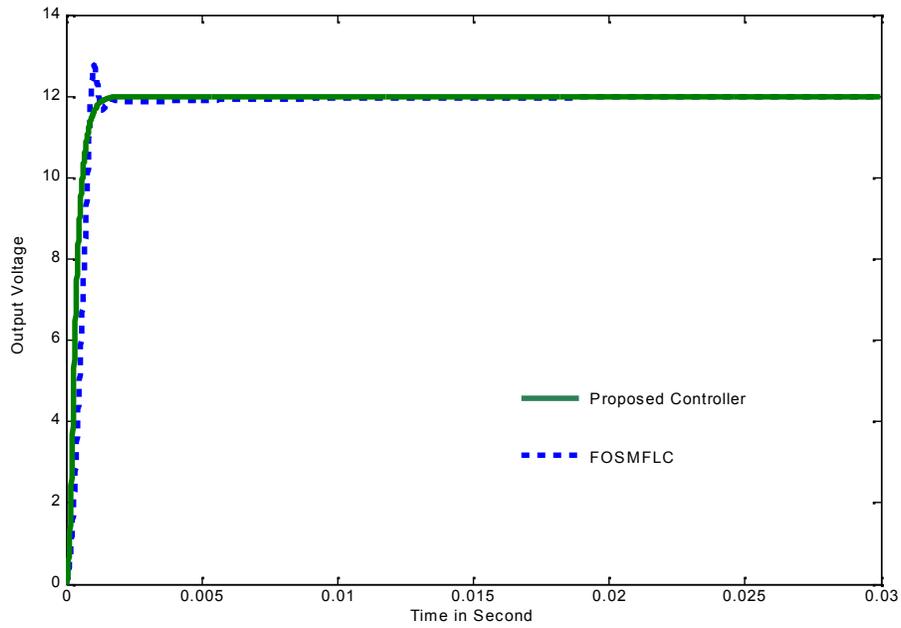

Fig 9. The Proposed Controller and FOSMFLC responses when the load resistance changes from 10Ω to 5Ω.

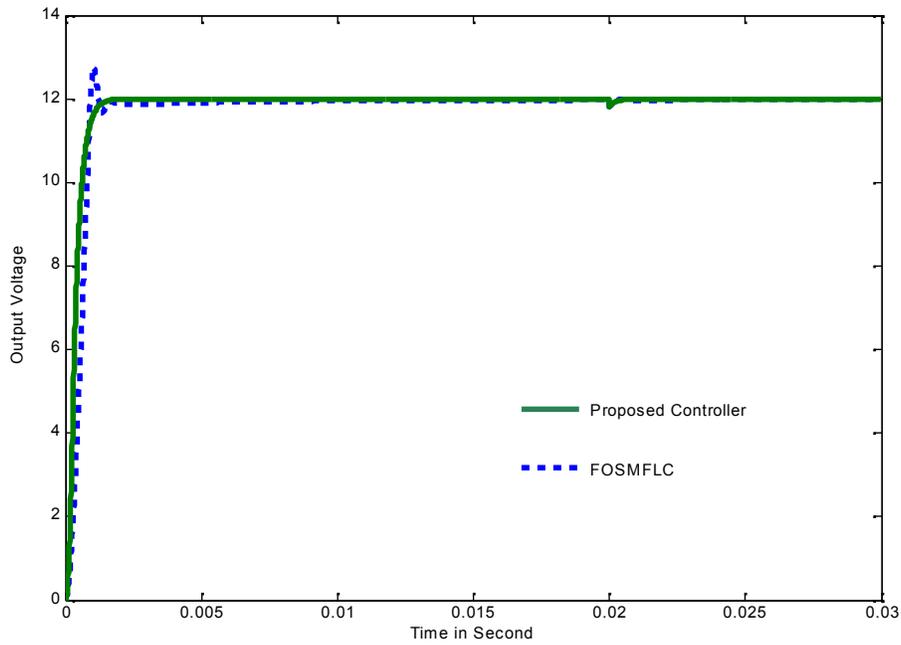

Fig 10. The Proposed Controller and FOSMFLC responses when the load resistance changes from 10Ω to 100Ω.

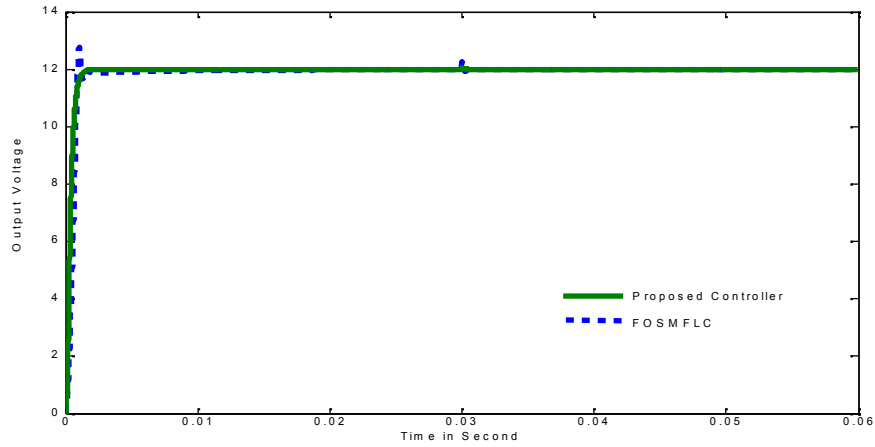

Fig 11. The Proposed Controller and FOSMFLC responses when the load resistance changes from 10Ω to 1Ω.

The Proposed controller gives good start up transient response for different input voltages. The responses are almost similar for different input voltages. The proposed controller gives faster start-up transient response and less overshoot compared to first order Sliding Mode Fuzzy logic Controller. Also, the proposed controller is able to respond quickly to input voltage variations and load variations at steady-state. The simulation results are tabulated in Table 3 to Table 5.

|  | Proposed Controller | FOSMFLC |
|---|---|---|
| Settling time (5 % of Steady-State value) | 0.65 ms | 1.58 ms |
| Overshoot | 0 | 8.5% |
| Steady-State error | 0 | 0 |

Table 3. Comparison of the responses of DC-DC Buck Converter using the Proposed Controller and FOSMFLC.

|  |  | Proposed Controller | FOSMFLC |
|---|---|---|---|
| When input voltage is 22V | Settling time | 0.78 ms | 1.54 ms |
|  | Overshoot | 2.75 % | 11.33 % |
|  | Steady-State error | 0 | 0 |
| When input voltage is 18V | Settling time | 0.72 ms | 1.64 ms |
|  | Overshoot | 0 | 6.1 % |
|  | Steady-State error | 0 | 0 |

Table 4. Comparison of the Start-up transient responses of DC-DC Buck Converter using the Proposed Controller and FOSMFLC for different input voltages.

|  |  | Proposed Controller | SOMFLC |
|---|---|---|---|
| When input Voltage increases by 2 V | Time taken to reject Input Voltage Variation | 1.2 ms | 3 ms |
|  | Steady-State error | 0 | 0 |
| When input Voltage decreases by 2 V | Time taken to reject Input Voltage Variation | 1.4 ms | 12 ms |
|  | Steady-State error | 0 | 0 |

Table 5. Comparison of the responses of DC-DC Buck Converter using the proposed controller and FOSMFLC for different variations in input voltages at Steady-State.

# 7. CONCLUSION

A novel fuzzy logic based Adaptive Super-twisting Sliding Mode Controller is proposed for the control of dynamic uncertain systems. The proposed controller combines the advantages of Second order Sliding Mode Control, Fuzzy Logic Control and Adaptive Control. The reaching conditions, stability and robustness of the system with the proposed controller are guaranteed. In addition, the proposed controller is well suited for the simple design and implementation. The effectiveness of the proposed controller over the first order Sliding Mode Fuzzy Logic controller is illustrated by Matlab/Simulink based simulations performed on a DC-DC Buck converter. Based on this comparison, the proposed controller is shown to obtain the desired transient response without causing chattering and error under steady-state conditions. The proposed controller is able to give robust performance in terms of rejection to input voltage variations and load variations.

## Authors

**Abdul Kareem** graduated with B.Tech in Electrical and Electronics Engineering from LBS College of Engineering, Kasaragod and M.Tech in Microelectronics and Control systems from VTU Extension Centre, NMAMIT, NITTE. Currently, he is pursuing Ph. D under the guidance of Dr. Mohammad Fazle Azeem in St. Peter's University, Chennai, India. He is an Associate Professor in the department of Electronics and Communication Engineering, Sahyadri College of Engineering and Management, Mangalore, India. He has served as a resource person in many conferences and workshops on Soft Computing. His research interests include Soft Computing, Intelligent Control systems.

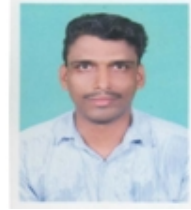

**Dr. Mohammad Fazle Azeem** completed his BE in Electrical Engineering from MMMEC, Gorakhpur and M.Tech in Electrical Engineering from Aligarh Muslim University, Aligarh and Ph. D from IIT, Delhi. He is working as a Professor in the department of Electronics and Communication Engineering, PA College of Engineering, Mangalore, India. He has to his credit 55 research publications. He is a well known resource person in the field of Soft Computing and has delivered talks in many workshops and Conferences. His research interests include Soft Computing, Non-linear Control, Signal Processing, Low Power VLSI.

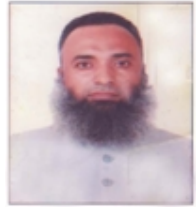